\documentclass[10pt,twocolumn,letterpaper]{article}

\usepackage{cvpr}      
\usepackage{graphicx}
\usepackage{amsmath}
\usepackage{amssymb}
\usepackage{booktabs}

\usepackage{times}
\usepackage{epsfig}
\usepackage{graphicx}
\usepackage{amsmath}
\usepackage{amssymb}
\usepackage{times}
\usepackage{epsfig}
\usepackage{graphicx}
\usepackage{amsmath}
\usepackage{amssymb}
\usepackage{listings}
\usepackage{algorithm}
\usepackage{graphicx}
\usepackage{booktabs}
\usepackage{stmaryrd}

\usepackage[dvipsnames]{xcolor}
\usepackage{multirow}

\definecolor{dkgreen}{rgb}{0,0.6,0}
\definecolor{gray}{rgb}{0.5,0.5,0.5}
\definecolor{mauve}{rgb}{0.58,0,0.82}

\usepackage{algorithm}
\usepackage{algorithmic}

\usepackage{newfloat}
\usepackage{listings}
\floatstyle{ruled}
\newfloat{listing}{tb}{lst}{}
\floatname{listing}{Listing}

\usepackage[pagebackref,breaklinks,colorlinks]{hyperref}

\usepackage[capitalize]{cleveref}
\crefname{section}{Sec.}{Secs.}
\Crefname{section}{Section}{Sections}
\Crefname{table}{Table}{Tables}
\crefname{table}{Tab.}{Tabs.}

\def\cvprPaperID{7399} 
\def\confName{CVPR}
\def\confYear{2022}

\begin{document}

\title{End-to-End Segmentation via Patch-wise Polygons Prediction}

\author{Tal Shaharabany\\
Tel-Aviv University\\
{\tt\small shaharabany@mail.tau.ac.il}
\and
Lior Wolf\\
Tel-Aviv University\\
{\tt\small liorwolf@gmail.com}
}

\maketitle

\begin{abstract}
The leading segmentation methods represent the output map as a pixel grid. We study an alternative representation in which the object edges are modeled, per image patch, as a polygon with $k$ vertices that is coupled with per-patch label probabilities. The vertices are optimized by employing a differentiable neural renderer to create a raster image. The delineated region is then compared with the ground truth segmentation. Our method obtains multiple state-of-the-art results: 76.26\% mIoU on the Cityscapes validation, 90.92\% IoU on the Vaihingen building segmentation benchmark, 66.82\% IoU for the MoNU microscopy  dataset, and 90.91\% for the bird benchmark CUB. Our code for training and reproducing these results is attached as supplementary.
\end{abstract}

\section{Introduction}

Accurate segmentation in dense environments is a challenging computer vision task. Many applications, such as autonomous driving~\cite{treml2016speeding}, drone navigation~\cite{parmar2020exploration}, and human-robot interaction~\cite{oliveira2018efficient}, rely on segmentation for performing scene understanding.  Therefore, the accuracy of the segmentation method should be high as possible.

In recent years, fully convolutional networks based on encoder-decoder architectures~\cite{long2015fully}, in which the encoder is pre-trained on ImageNet, have become the standard tool in the field. Several techniques were developed to better leverage the capacity of the architectures. For example, skip connections between the encoder and the decoder were added to overcome the loss of spatial information and to better propagate the training signal through the network~\cite{ronneberger2015u}. 
What all these methods have in common is the representation of the output segmentation map as a binary multi-channel image, where the number of channels is the number of classes $C$. In our work, we add a second type of output representation, in which, for every patch in the image and for every class, a polygon comprised of $k$ points represents the binary mask.

\begin{figure*}[t]
\centering
    \begin{tabular}{ccc}
    \includegraphics[scale=0.55]{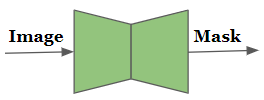}&
     \multirow{3}{*}{\raisebox{-.782in}[0pt][0pt]{\includegraphics[scale=0.4]{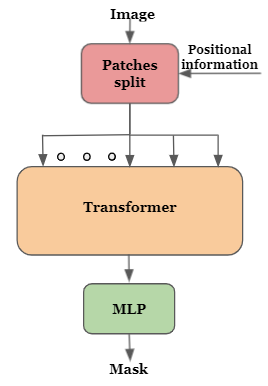}}} 
 & \multirow{3}{*}{\raisebox{-.6in}[0pt][0pt]{\includegraphics[scale=0.5]{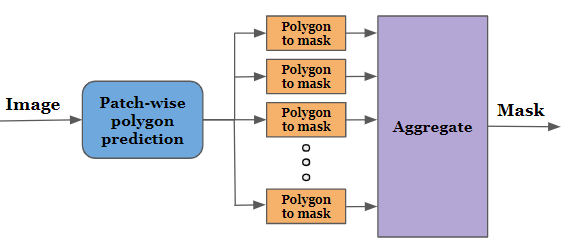}}}\\ (a)\\
    {\includegraphics[scale=1.0]{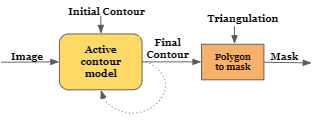}} & \\
    (b) & (c) & (d)\\

    \end{tabular}
\caption{Comparison of the proposed method to the common segmentation methods. (a) The prevalent encoder-decoder architecture predicts a pixel-wise segmentation mask. (b) Deep active contour approaches, in which the trained network predicts the contour of the object as a single polygon. The vertices of the polygon are updated iteratively. (c) Transformer-based segmentation models first divide the input image into per-patch tokens, each associated with a positional encoding. The transformer algorithm then adds multiple layers of interactions between the tokens. An MLP model then generates a local mask per each token. (d) In our solution, the network predicts a single polygon for every patch in the image, in addition to low-resolution probability maps (not shown). Each polygon is rendered into a local mask segmentation mask, and all the masks are aggregated to obtain the final segmentation mask.} 
    \label{fig:teaser}
\end{figure*}

Alternative image segmentation architectures are presented in Fig.~\ref{fig:teaser}. The most common is the encoder-decoder one. The active contour methods, similar to our work, employ a polygon representation. Such methods, however, are iterative, and produce a single polygon, lacking the flexibility to cope with occlusions. 
Transformer based network employ patch-wise self-attention maps to calculate a spatial representation that attends globally to multiple parts of the image. The output mask is then generated by an MLP decoder. Our method differs from these approaches and produces multiple polygons, which are rendered by a neural renderer to produce the final segmentation mask.

For optimization purposes, we use a neural renderer to translate the polygons to a binary raster-graphics mask. The neural renderer provides a gradient for the polygon vertices and enables us to maximize the overlap between the output masks and the ground truth. 
Since there are polygons for every patch and every class, whether the object appears there or not, we multiply the binary mask with a low-resolution binary map that indicates the existence of each object.

The method achieves state-of-the-art results on four segmentation benchmarks, Cityscapes,  CUB-200-2011, Nucleus segmentation challenges (MoNuSeg) and on the Vaihingen building segmentation benchmark. 

\section{Related work}
Most modern approaches to image segmentation perform image-to-image mapping with a Fully Convolutional Network~\cite{long2015fully}. Many encoder-decoder architectures were proposed, e.g.,~\cite{he2016deep, szegedy2015going, simonyan2014very, huang2017densely}, most of which employ U-Net based skip connections between parallel blocks of the encoder and the decoder~\cite{ronneberger2015u}.  Improved generalization is obtained by reducing the gap between the encoder and decoder semantic maps~\cite{zhou2018unet++}. Another contribution adds residual connections for each encoder and decoder block and divides the image into patches with a calculated weighted map for each patch as input to the model.~\cite{xiao2018weighted}.  Wang et al. add attention maps to each feature map in the encoder-decoder block in order to enlarge the representation between far areas in the image with good computational efficiency \cite{wang2020axial}.  This is extended by adding scalar gates to the attention layers~\cite{valanarasu2021medical}. The same contribution also adds a Local-Global training strategy (LoGo), which further improves performance. Dilated convolutions~\cite{yu2015multi} are used by many methods to preserve the spatial size of the feature map and to enlarge the receptive field~\cite{chen2014semantic,  zhao2017pyramid, zhang2018context}. The receptive field can also be enlarged by using larger kernels~\cite{peng2017large} and enriched with contextual information, by using kernels of multiple scales~\cite{chen2017deeplab}. Hardnet~\cite{chao2019hardnet} employs a Harmonic Densely Connected Network that is shown to be highly effective in many real-time tasks, such as classification and object detection. 

\noindent{\bf Polygon-based segmentation\quad} Active contour, also known as snakes~\cite{kass1988snakes}, was used on many instance segmentation applications over the years. Such methods iteratively minimize an energy function that includes both edge-seeking terms and a shape prior until it stops~\cite{kichenassamy1995gradient, cohen1991active, caselles1997geodesic}. The first deep learning-based snakes predict a correction vector for each point, by considering the small enclosing patch, in order to converge to the object outline~\cite{rupprecht2016deep}.   Acuna et al.\cite{acuna2018efficient} propose automated and semi-automated methods for object annotating and segmentation based on Graph neural networks and reinforcement learning. Ling et al.\cite{ling2019fast} used a Graph Convolutional Network (GCN) in order to fit the polygon's positions to the outline. A different approach predicts a shift map and iteratively changes the positions of the polygon vertices, according to this map~\cite{gur2019end}. In order to obtain a gradient signal, the authors use a neural renderer that transforms the 2D polygon into a 2D segmentation map. {Our method also employs a neural renderer. However, we perform local predictions per patch and recover the vertex locations of the polygon in each patch non-iteratively. This allows us to perform finer local predictions, and overcome discontinuities caused by occlusions or by the existence of multiple objects.}  

A recent contribution~\cite{liang2020polytransform}, refines the output instance mask of pretrained Mask-RCNN algorithm by learning to correct the coordinates of the extracted polygons. This algorithm obtains good performances for the cityscapes instance segmentation benchmark but uses different settings than the rest of the literature. For example, the entire image is used instead of a specified bounding box, and the bounding box expansion differs from what is conventionally used in published work. Since no code was published, we cannot perform a direct comparison. From the algorithmic perspective, the methods greatly differ in the type of polygons used (boundary vs. local), the type of optimization (Chamfer Distance loss~\cite{homayounfar2018hierarchical} on polygon coordinates vs. using a neural renderer), and almost every other aspect.

\noindent{\bf Neural Renderers\quad} The ground truth segmentation is given as an image, and in order to compare the obtained solution with it and compute a loss term, it is required to have a differentiable way to convert the vertex representation to a rasterized form. This conversion task, when implemented as a neural network, is referred to as ``neural renderers''. On the backward pass, fully differential neural renderers convert image-based loss to gradients on the vertices' coordinates. Loper et al.~\cite{loper2014opendr} approximate this gradient-based on image derivatives, while Kato et al.~\cite{kato2018neural} approximates the propagated error, according to the overlap between the rasterization output and the ground truth.

\begin{figure*}[t]
\centering
    \begin{tabular}{ccc}
    \includegraphics[width=0.255\linewidth]{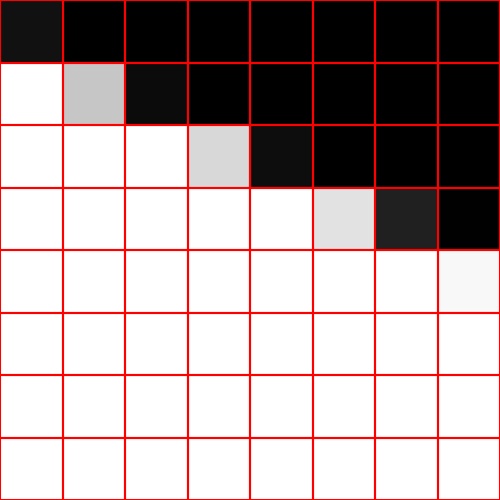} &
    \includegraphics[width=0.255\linewidth]{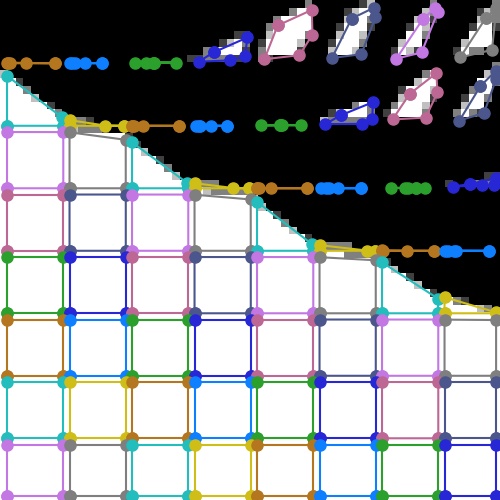} &
    \includegraphics[width=0.255\linewidth]{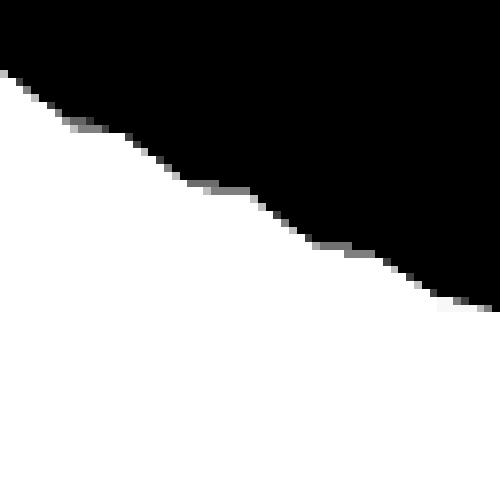} \\
    
    (a)&(b)&(c)\\
    \end{tabular}

\caption{Example of the process of combining the polygon data with the grid-based predictions, in order to obtain a refined segmentation map. (a) The low-resolution segmentation map  $M$ is shown for a single object (zoom-in from the Cityscapes dataset). White pixels denote in-object, blacks are for the background. (b) The corresponding polygons $R_p$. Despite using $k=5$, many polygons match the patch boundaries. Different polygons are drawn in different colors and are slightly scaled to visualize them separately. (c) The combined segmentation map $M_o$, in the final resolution. The polygons on the top right part of $R_p$ are gated out by $M$.} 
    \label{fig:poly}
\end{figure*}

\section{Methods}

The input to our method is an RGB image $I$, of dimensions ${3 \times  H \times W}$. The image is divided into patches of size $s\times s$ and a polygon with $k$ vertices is predicted for each patch. In addition, each patch has an associate scalar that determines the probability of the object being present in the patch. 

This mask representation is illustrated in Fig.~\ref{fig:poly}. The low-res probability map in panel (a) assigns one value per grid cell, where we denote higher probabilities by a brighter value. The local polygons in panel (b) determine a local segmentation mask in each grid cell. When the grid cell is completely enclosed by the segmented object, the polygon covers the entire cell. At object boundaries, it takes the form of the local boundary. When out of the object, it can degenerate to a zero area polygon or an arbitrary one. The final segmentation map, depicted in Fig.~\ref{fig:poly}(c) is obtained by multiplying the low-resolution probability map with the local patch rastered for each polygon.

The learned network we employ is given as $f=f_2\circ f_1$, where $f_1: \mathbb{R}^{C \times  {H} \times {W}} \rightarrow \mathbb{R}^{C \times  \frac{H}{32} \times \frac{W}{32}}$ is the backbone (encoder) network, and $f_2$ is the decoder networks. The output space of $f_2$ is a tensor that can be viewed as a field with $\frac{H}{s} \times \frac{W}{s}$ vectors $\mathbb{R}^{2k+1}$, each representing a single polygon that is associated with a single image patch. This representation is the concatenation of the 2D coordinates of the $k$ vertices together with the probability. The image coordinates are normalized, such that their range is [-1,1]. 

\begin{figure}[t]
    \centering
    \begin{tabular}{c}
    \includegraphics[width=1\linewidth]{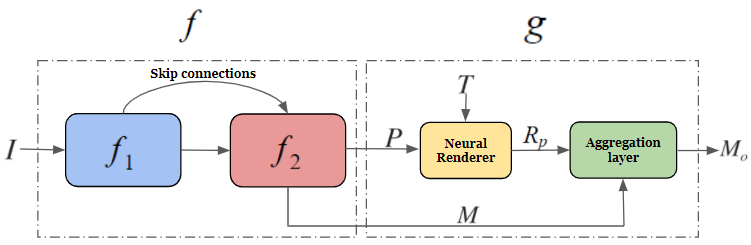} 
    \end{tabular}
    \caption{The proposed method. Network $f$ contains an encoder-decoder network with skip connections between them. Given an input Image I, it generates (1) a low-resolution segmentation probability map $M$ (2) a patch-wise polygon array $P$. Network $g$ contains (1) a neural renderer that employs a predefined triangulation $T$ to generate the binary polygon segmentation map $R_p$ from $P$ (2) an aggregation layer that rearranges $R_p$  according to the patch location in the original image. The accumulated map is then gated, i.e., multiplied pixel-wise, by the initial probabilities provided by the upsampled version of $M$ to generate the output segmentation map $M_o$.}
    \label{fig:arch}
\end{figure}

A fixed triangulation $T$ is obtained by performing Delaunay triangulation to a regular polygon of $k$ vertices. It is used to convert all polygons to a raster image, assuming that locally the polygons are almost always close to being convex. We set the z-axis to be constant and the perspective to be zero, in order to use a 3D renderer for 2D objects.

The output $f(I)$ is divided into two parts, see Fig.~\ref{fig:arch}. We denote by $P=\{P^i\}$ the set of all $\frac{H}{s} \times \frac{W}{s}$ polygons, and by $M$ the map of size $\frac{H}{s} \times \frac{W}{s}$ that contains all the probabilities. The neural renderer $r$, given the vertices $P$ and the fixed triangulation $T$,  returns the polygon shape as a local image patch. In this raster image, all pixels inside the polygon defined by $P$ through the triangulation $T$ are assigned a value of one, zero for pixels outside, and in-between values for pixels that the polygon divides:
\begin{equation}
   {M^{i}_p} = r(P^i,T) \in \mathbb{R}^{s\times s}\,,
\end{equation}
The aggregation layer collecting all local patches $M^{i}_p$, each in its associated patch location, we obtain the map $R_p\in \mathbb{R}^{H \times W}$. By performing nearest neighbor upsampling to the probability map $M$ by a factor of $s$ in each dimension, and obtain a probability maps of the same size as $R_p$. The final mask is the pixel-wise product between these two maps.
\begin{equation}
   {M}_o = {\uparrow}M \odot {R_p}\,,
\end{equation}
where $\odot$ denotes elementwise multiplication and $\uparrow$ denotes upsampling. This way, the low-resolution mask serves as a gating mechanism that determines whether the polygon exists within an object of a certain class in the output segmentation map.

To maximize the overlap between the output mask  (before and after incorporating the polygons) and the ground truth segmentation map $Y$, a soft dice loss \cite{sudre2017generalised}:
\begin{equation}
\mathcal{L}_{dice}(y,\hat{y}) = 1 - \frac{2TP(y,\hat{y}) + 1}{2TP(y,\hat{y}) + FN(y,\hat{y}) + FP(y,\hat{y}) + 1 }
\end{equation}
Where TP is the true positive between the ground truth $y$ and output mask $\hat{y}$, FN is a false negative and FP is a false positive.
In addition, a binary cross-entropy is used for $M$:
\begin{equation}
\mathcal{L}_{BCE}(y, \hat{y}) = -y\log(\hat{y}) + (1-y)\log(1-\hat{y})
\end{equation}

\begin{equation}
\mathcal{L}  = \mathcal{L}_{BCE}({\uparrow}M, Y) + \mathcal{L}_{dice}(M_o, Y)
\end{equation}
This loss appears without weighting, in order to avoid an additional parameter. During inference, we employ only $M_o$ and the second term can be seen as auxiliary.

\paragraph{Architecture}
For the backbone $f_1$, the Harmonic Dense Net \cite{chao2019hardnet} is used, in which the receptive field is of size $32\times 32$. The network contains six ``HarD'' blocks, as described in \cite{chao2019hardnet}, with 192, 256, 320, 480, 720, 1280 output channels, respectively. The last block contains a dropout layer of 0.1. Pre-trained ImageNet weights are used at initialization.

The decoder $f_2$ contains two upsampling blocks in order to obtain an output resolution of $s=8$ times less than the original image size.  Each block contains two convolutional layers with a kernel size equal to 3 and zero padding equal to one. In addition, we use batch normalization after the last convolution layer before the activation function. The first layer's activation function is a ReLU, while the second layer's activation functions are sigmoid for the channel of the low-resolution map and $tanh$ for the polygon vertices. Each layer receives a skip connection from the block of the encoder that has the same spatial resolution. We note that our decoder requires considerably fewer learnable parameters related to a regular decoder and, since only two blocks are used, fewer skip connections from the encoder.   

The subsequent network $g$ employs the mesh renderer of \cite{kato2018neural}, with a zero perspective, a camera distance of one, and output mask of size $s \times {s}$.  This way, the 3D renderer is adapted for the 2D rendering task.

In term of FLOPs, for image size of $256^2$ we get 16.51 GMacs for our model and 25.44 GMacs for the comparable FCN. The peak memory consumption is 486MB for our model and 506MB for the FCN.

\begin{table}[t]
\centering
\begin{tabular}{@{}lcc@{}}
\toprule
Category & Full[train/val] & Occlusion[train/val] \\
\midrule
Bicycle & 3501/1166 & 1029/385\\
Bus & 374/98 & 115/30\\
Car & 26226/4655 & 3333/914\\
Person & 16753/3395 & 1731/669\\
Train & 167/23 & 86/15\\
Truck & 481/93 & 108/19\\
Motorcycle & 707/149 & 175/55\\
Rider & 1711/544 & 291/155\\
\bottomrule
\end{tabular}
\caption{Cityscapes partitions training and validation set for each category, for the full dataset, and the occlusion subset.}
\label{tab:city_data}
\end{table}

\begin{figure*}[t]
    \centering
    \setlength{\tabcolsep}{4.5pt} 
    \renewcommand{\arraystretch}{2} 
    \begin{tabular}{cccc}

    \includegraphics[width=0.18\linewidth]{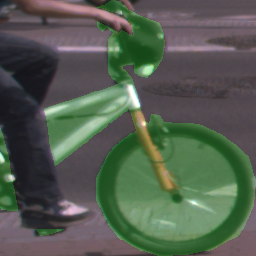} &
    \includegraphics[width=0.18\linewidth]{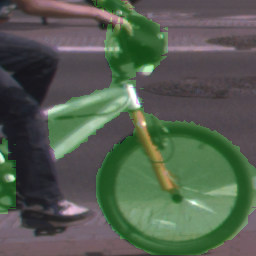} &
    \includegraphics[width=0.18\linewidth]{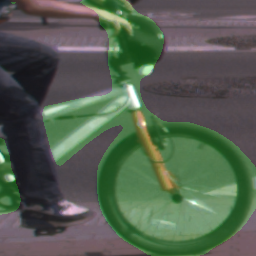} &
    \includegraphics[width=0.18\linewidth]{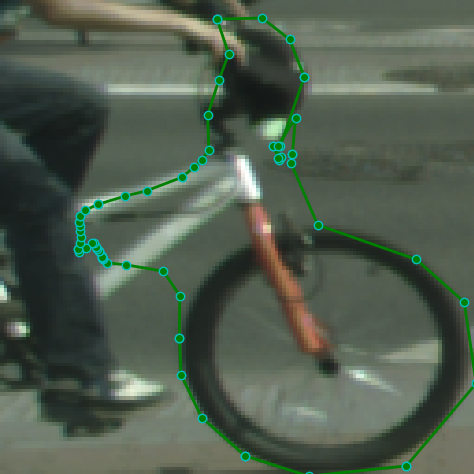} \\
    \includegraphics[width=0.18\linewidth]{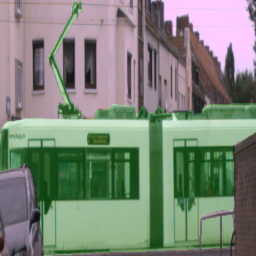} &
     \includegraphics[width=0.18\linewidth]{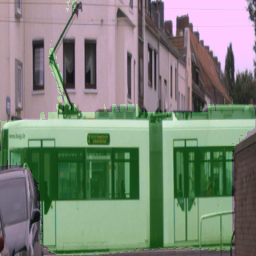} &
      \includegraphics[width=0.18\linewidth]{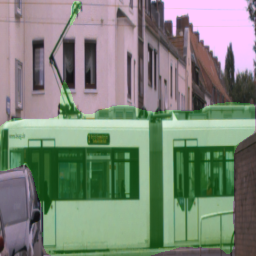} &
      \includegraphics[width=0.18\linewidth]{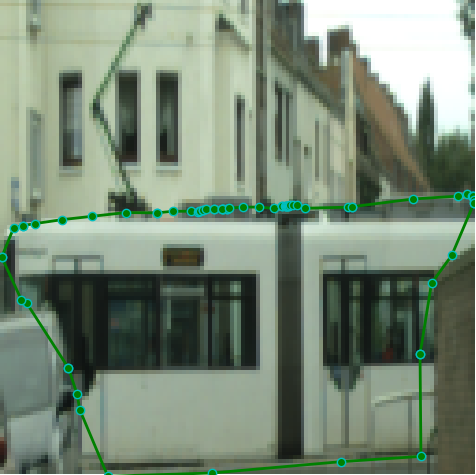} \\
      
    (a) & (b) & (c) & (d) 
    \end{tabular}
    \caption{Sample segmentation masks on the Cityscapes validation dataset. (a) ground-truth (b) our method (c) FCN-HarDNet-85~\cite{chao2019hardnet} (d) the result of the active contour approach of~\cite{gur2019end}. The active contour method cannot handle occlusions or objects with overhanging parts.} 
    \label{fig:citysample}
\end{figure*}

\begin{table*}[t]
\centering
\begin{tabular}{@{~}lccccccccc@{~}}
\toprule
Method & Bike & Bus & Person & Train & Truck & Mcycle & Car & Rider & Mean \\
\midrule
Polygon-RNN++ (with BS) \cite{acuna2018efficient} & 63.06 &  81.38 & 72.41 & 64.28 & 78.90 & 62.01 & 79.08 & 69.95 & 71.38\\
PSP-DeepLab  \cite{chen2017deeplab} & 67.18 &  83.81 & 72.62 & 68.76 & 80.48 & 65.94 & 80.45 & 70.00 & 73.66\\
Polygon-GCN (with PS)  \cite{ling2019fast} & 66.55 &  85.01 & 72.94 & 60.99 & 79.78 & 63.87 & 81.09 & 71.00 & 72.66\\
Spline-GCN (with PS)  \cite{ling2019fast} & 67.36 &  {\bf85.43} & 73.72 & 64.40 & 80.22 & 64.86 & 81.88 & 71.73 & 73.70\\
Ours & {\bf 68.20} & 85.10 & {\bf 74.52} & {\bf 74.80} & {\bf 80.50} & {\bf 66.20} & {\bf 82.10} & {\bf 71.81} & {\bf 75.40}\\
\midrule
\midrule
Deep contour \cite{gur2019end} & 68.08 &  83.02 & 75.04 & 74.53 & 79.55 & 66.53 & 81.92 & 72.03 & 75.09\\
FCN-HarDNet-85 \cite{chao2019hardnet} & 68.26 &  84.98 & 74.51 & 76.60 & 80.20 & 66.25 & 82.36 & 71.57 & 75.59\\
Ours & {\bf 69.53} & {\bf 85.50} & {\bf 75.15} & {\bf 76.90} & {\bf 81.20} & {\bf 66.96} & {\bf 82.69} & {\bf 72.18} & {\bf 76.26}\\
\bottomrule
\end{tabular}
\caption{Cityscapes segmentation results for two protocols: the top part refers to segmentation results with 15\% expansion around the bounding box; the bottom part refers to segmentation results with no expansion around the bounding box}
\label{tab:cityscapes}
\end{table*}

\begin{table}[t]
\centering
\begin{tabular}{@{}l@{~}c@{~}c@{~}c@{~}c@{}}
\toprule
Method & F1-Score & mIoU & WCov & FBound \\
\midrule
FCN-UNet~(Ronneberger 2015) & 87.40 & 78.60 & 81.80 & 40.20\\
FCN-ResNet34 & 91.76 & 87.20 & 88.55 & 75.12\\
FCN-HarDNet-85 (Chao 2019) & 93.97 & 88.95 & 93.60 & 80.20\\
DSAC (Marcos 2018) & - & 71.10 & 70.70 & 36.40  \\
DarNet (Cheng 2019) & 93.66 & 88.20 & 88.10 & 75.90\\
Deep contour (Gur 2019) & 94.80 & 90.33 & 93.72 & 78.72\\
TDAC (Hatamizadeh 2020) & 94.26 & 89.16 & 90.54 & 78.12\\
Ours  & {\bf 95.15} & {\bf 90.92} & {\bf 94.36} & {\bf 83.89}\\
\bottomrule
\end{tabular}
\caption{Segmentation results on the Vaihingen building dataset.}
\label{tab:building}
\end{table}

\begin{figure}[t]
\centering   \includegraphics[width=0.9\linewidth]{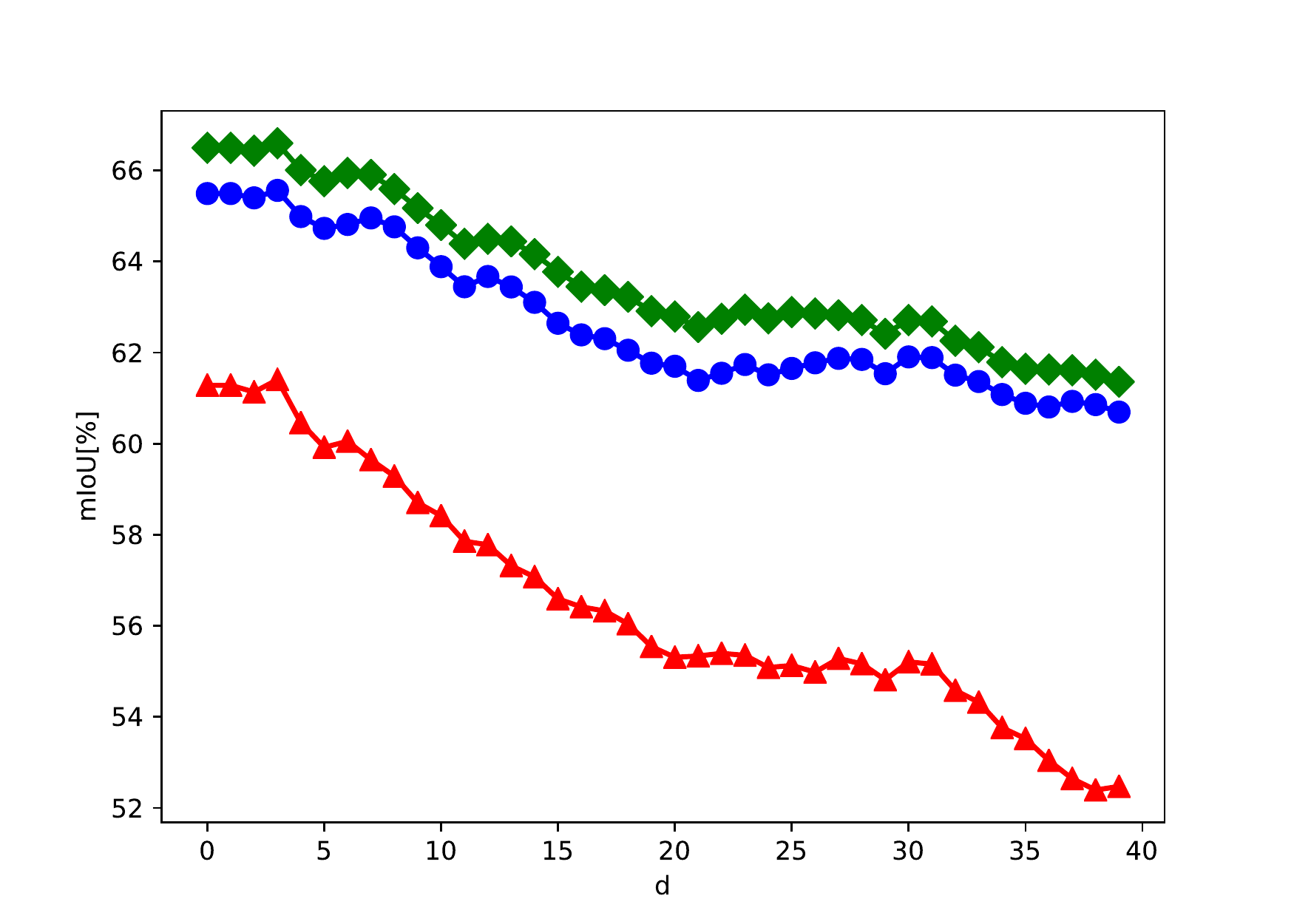}
    \caption{Performance under occlusion of our method (green diamond), in comparison to the leading UNet like network FCN-Hardnet85~\cite{chao2019hardnet} (blue circle) and the state-of-the-art active contour method~\cite{gur2019end} (red triangle). The x-axis denotes the amount of occlusion (see text). Shows is mean across all classes, see supplementary for a breakdown per class.}
    \label{fig:Occlusions}
\end{figure}

\begin{figure*}[t]
    \setlength{\tabcolsep}{4.5pt} 
    \renewcommand{\arraystretch}{2} 
    \centering
    \begin{tabular}{cccc}
    \includegraphics[width=0.19\linewidth]{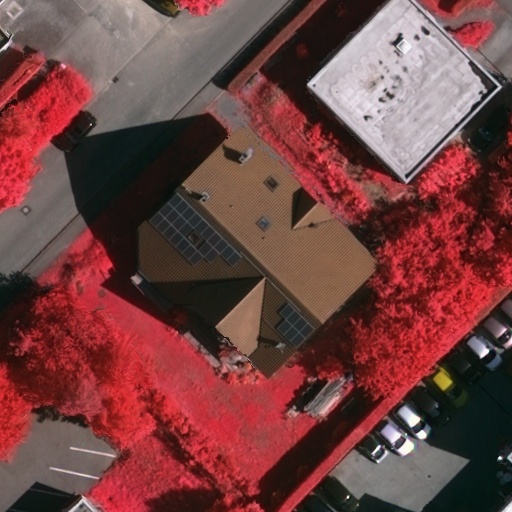} &
    \includegraphics[width=0.19\linewidth]{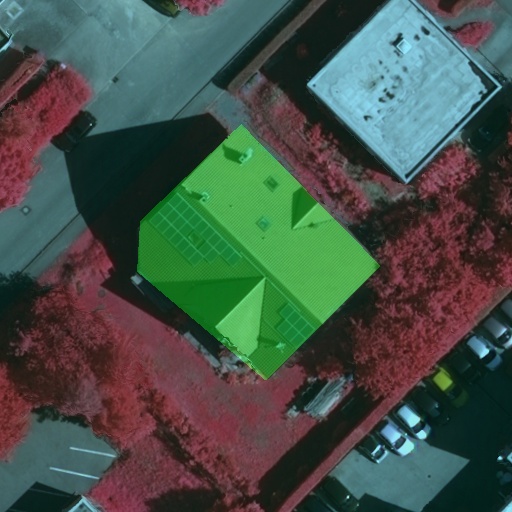} &
    \includegraphics[width=0.19\linewidth]{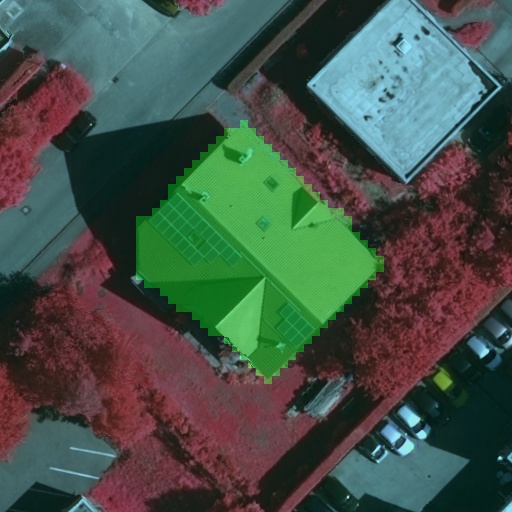} &
    \includegraphics[width=0.19\linewidth]{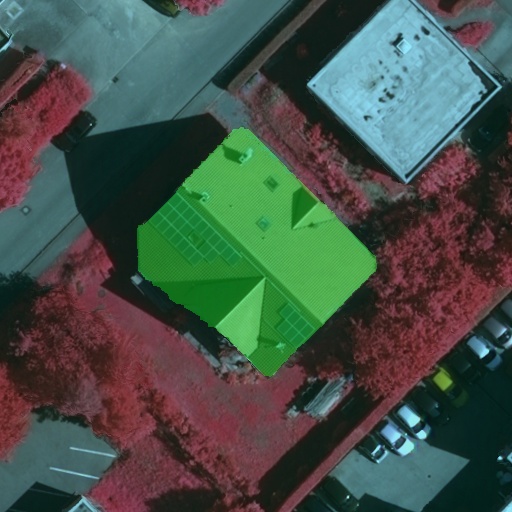} \\
    \includegraphics[width=0.19\linewidth]{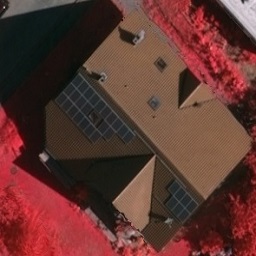} &
    \includegraphics[width=0.19\linewidth]{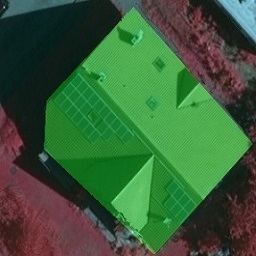} &
    \includegraphics[width=0.19\linewidth]{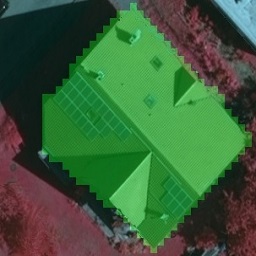} &
    \includegraphics[width=0.19\linewidth]{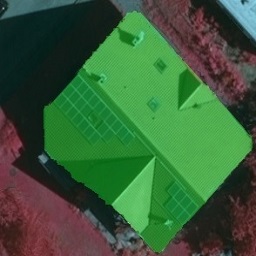} \\
\vspace{-1mm}
    (a) & (b) & (c) & (d)
    \end{tabular}
    \caption{Sample results of the proposed method on the building dataset. (a) The input image. (b) Ground truth segmentation. (c) The initial segmentation map $M$. (d) The final segmentation map $M_o$. First row: original images; second: a zoom-in view.} 
    \label{fig:buildingsample}
\end{figure*}

\begin{table}[t]
\centering
\begin{tabular}{lcc}
\toprule
Method & Dice & mIoU \\
\midrule
FCN \cite{badrinarayanan2017segnet} & 28.84 &  28.71 \\
U-Net \cite{ronneberger2015u} & 79.43 & 65.99 \\
U-Net++\cite{zhou2018unet++} & 79.49 & 66.04 \\
Res-UNet\cite{xiao2018weighted} & 79.49 & 66.07 \\
Axial Attention U-Net\cite{wang2020axial} & 76.83 & 62.49 \\
MedT\cite{valanarasu2021medical} & 79.55 & 66.17 \\
\midrule
FCN-Hardnet85 & 79.52 & 66.06\\
Low res FCN-Hardnet85 ($M$)  & 65.82 & 49.13 \\
Ours ($M_o$) & {\bf80.05} & {\bf66.82} \\
\bottomrule
\end{tabular}
\caption{Results for MoNu dataset}
\label{tab:Monu}
\end{table}

\begin{figure}[t]
    \setlength{\tabcolsep}{4.5pt} 
    \renewcommand{\arraystretch}{2} 
    \centering
    \begin{tabular}{cccc}
    \includegraphics[width=0.22\linewidth]{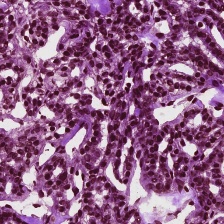} &
    \includegraphics[width=0.22\linewidth]{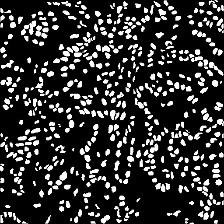} &
    \includegraphics[width=0.22\linewidth]{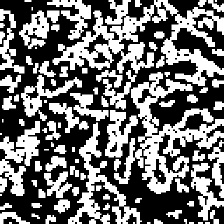} &
    \includegraphics[width=0.22\linewidth]{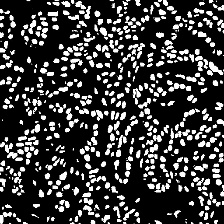} \\
    \includegraphics[width=0.22\linewidth]{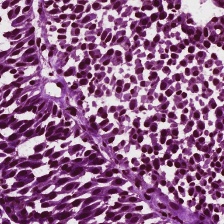} &
    \includegraphics[width=0.22\linewidth]{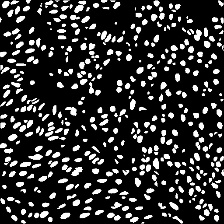} &
    \includegraphics[width=0.22\linewidth]{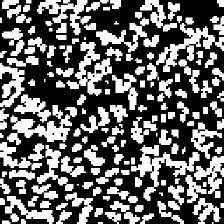} &
    \includegraphics[width=0.22\linewidth]{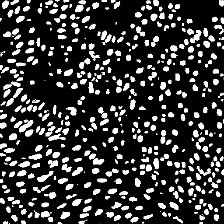} \\
    \includegraphics[width=0.22\linewidth]{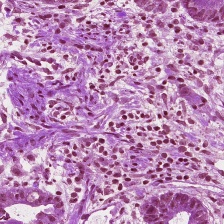} &
    \includegraphics[width=0.22\linewidth]{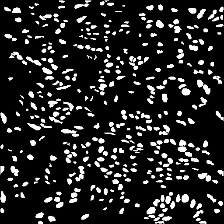} &
    \includegraphics[width=0.22\linewidth]{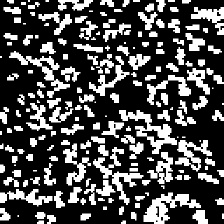} &
    \includegraphics[width=0.22\linewidth]{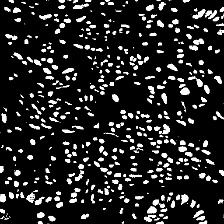} \\
\vspace{-1mm}
    (a) & (b) & (c) & (d)
    \end{tabular}
    \caption{Sample results of the proposed method on the MoNuSeg dataset. (a) The input image. (b) Ground truth segmentation. (c) The low-resolution segmentation map $M$. (d) The final segmentation map $M_o$.}
    \label{fig:Monu_seg}
\end{figure}

\section{Experiments}
We present our results on three segmentation benchmarks and compare our performance with the state-of-the-art methods. We further perform a study on the effect of each parameter of the proposed algorithm, and on the effect of the underlying network architecture. We put an added emphasis on the subset of test images that depict an occlusion of the segmented object.

The Citycapes dataset~\cite{cordts2016cityscapes} for instance segmentation includes eight object classes with $5,000$ fine-annotated images, divided into $2,975$ images for training, $500$ for validation, and $1,525$ for the test set. 

The Cityscapes dataset contains urban scenes that are characterized by dense and multiple objects. In order to fully understand the scene, the segmentation method is required to be robust to occlusions of objects, which causes the segmentation task to be more challenging. Following~\cite{ling2019fast}, we train and evaluate given the bounding box of the object, such that the network outputs a single mask for the entire instance.

To analyze the images with occlusions, the cases where the ground truth contains at least two contours are identified. for this purpose, we define an occlusion distance $d$ per image as:
\begin{equation}
d = \max_{C_{i},C_{j} \in C}   \min_{p\in C_i, q\in C_j} \|p-q\|\,,
\end{equation}
where $C$ is the set of all contours of a specific class defined for a specific image, each contour being a set of points. The distance $d$, which degenerates to zero if there is only one contour, is then used to stratify the test images by the number of occlusions. Tab.~\ref{tab:city_data} contains the number of images in both the validation set of the entire Cityscapes dataset, as well as for the subset of which the occlusion distance $d$ is larger than one. As can be seen, occlusions are fairly common for many of the objects.

The baseline methods we compare to on the Cityscapes dataset are PSPDeepLab\cite{chen2017deeplab},  Polygon-RNN++\cite{acuna2018efficient},  Curve-GCN\cite{ling2019fast} and Deep active contours \cite{gur2019end}. Another baseline, denoted as ``FCN-HarDNet-85'' employs a fully-convolutional network with the same backbone~\cite{chao2019hardnet} we used for our own method.

We note that our method employs 17M parameters in its decoder, while the ``FCN-HarDNet-85'' employs a decoder with 22M parameters. 

The Vaihingen~\cite{isprs} dataset consists of 168 aerial images, at a resolution of $256\times 256$, from this German city. The task is to segment the central building in a very dense and challenging environment, which includes other structures, streets, trees, and cars. The dataset is divided into 100 buildings for training, and the remaining 68 for testing.

For building segmentation, we use an encoder-decoder architecture based on the HarDNet-85~\cite{chao2019hardnet} backbone and produce $M$ at 1/8 resolution ($s=8$).

The baseline methods we compare to on the Vaihingen buildings dataset are ``DSAC''\cite{marcos2018learning}, ``DarNet''\cite{cheng2019darnet}, ``TDAC''\cite{hatamizadeh2020end} and Deep active contours \cite{gur2019end}. We also present results for ``FCN-UNET''~\cite{ronneberger2015u}, ``FCN-ResNet-34'' and ``FCN-HarDNet-85'', all based on fully-convolutional network decoders.

The MoNuSeg dataset~\cite{kumar2019multi} contains a training set with 30 microscopic images from seven organs with annotations of 21,623 individual nuclei, the test dataset contains 14 similar images. Following previous work,  we resized the images into a resolution of $512\times 512$~\cite{valanarasu2021medical}. An encoder-decoder architecture based on the HarDNet-85~\cite{chao2019hardnet} backbone. Since the recovered elements are very small, our method employs a map $M$ at 1/4 resolution ($s=4$), and not lower resolutions as in other datasets. 

The baseline methods we compare to on the Monu dataset are ``FCN''\cite{badrinarayanan2017segnet}, ``UNET''\cite{ronneberger2015u}, ``UNET++''\cite{zhou2018unet++}, Res-Unet\cite{xiao2018weighted}, Axial attention Unet\cite{wang2020axial} and Medical transformer \cite{valanarasu2021medical}.

Finally, we report results on CUB-200-2011 \cite{wah2011caltech}. The dataset contains 5994 images of different birds for training, and 5794 for validation. We trained the model for the same setting of cityscapes where the input image size is $256^2$ and the evaluation is on the original image size.

\noindent{\bf Training details\quad} For training our network, we use the ADAM optimizer with an initial learning rate of 0.0003 and a gamma decay of 0.7, applied every 50 epochs. The batch size is 32 and the parameter of the weight decay regularization is set to $5\cdot10^{-5}$. A single GeForce RTX 2080 Ti is used for training on all datasets. Our networks are trained for up to 1000 epochs.  

Following~\cite{gur2019end}, we used for the Cityscapes images a set of augmentations that includes: (i) scaling by a random scale factor in the range of $[0.75,1.25]$ and then cropping an image of size $256\times 256$, (ii) color jitter with the parameters of brightness sampled uniformly between $[0,0.4]$, a contrast in the range $[0,0.4]$, saturation in the range $[0,0.4]$, and hue in the range $[0,0.1]$, (iii) a random horizontal flip with a probability of 0.5. 

For the Vaihingen dataset, also following~\cite{gur2019end}, we used (i) a random rotation augmentation of 360 degrees and a scale range of $[0.75, 1.5]$, (ii) a random vertical and horizontal flip with a probability of 0.5, (iii) random color jitter with a maximal value of 0.6 for brightness, 0.5 for contrast, 0.4 for saturation, and 0.025 for hue. 

For the MoNu dataset, we used (i) a random rotation augmentation of \textpm20 degrees and a scale range of $[0.75, 1.25]$, (ii) a random horizontal flip with a probability of 0.5, (iii) random color jitter with a maximal value of 0.4 for brightness, 0.4 for contrast, 0.4 for saturation, and 0.1 for hue.

\noindent{\bf Evaluation Metrics\quad} The Cityscapes segmentation results employ the common metrics of mean Intersection-over-Union (IoU).  
The Vaihingen benchmark is equipped with a set of additional  evaluation metrics. The results are evaluation by the IoU of the single class, and the F1-score of retrieving building pixels.  The weighted Coverage (WCov) score is computed as the IoU for each object in the image weighted by the related size of the object from the whole image. 
Finally, the Boundary F-score (BoundF) is the average of the  F1-scores computed with thresholds ranging from 1 to 5 pixels around the ground truth boundaries, as described by~\cite{isprs}.

\noindent{\bf Results\quad} The Cityscapes results are presented in Tab.~\ref{tab:cityscapes}. As can be seen, our method outperforms all baseline methods, including active contour methods and encoder-decoder architectures. Furthermore, one can observe the gap in performance between our method and the U-Net-like method that is based on the same backbone architecture (``FCN-HarDNet-85''). This is despite the reduction in the number of trainable parameters and the size of the representation (each $8\times 8$ patch is represented by $k+1$ output floats, instead of 64 floats).

Fig.~\ref{fig:citysample} presents sample results and compares our method to the leading active contour method (``Deep contour'') and to the FCN-HarDNet-85 baseline. As can be seen, the active contour method struggles to model objects that deviate considerably from their convex hull, and cannot handle occlusions well. In comparison to the FCN baseline, our method provides more accurate details (consider the bicycle fork in the first row), but the overall shape is the same.

The results for images stratified by the level of occlusion (the measurement $d$) are presented in Fig.~\ref{fig:Occlusions}. 
All methods present a decrease in performance as the amount of occlusion increases. The FCN-HarDNet-85 maintains a stable performance level below our method, and the active contour method suffers the most from occlusion, as expected.

The results for the Vaihingen building dataset are presented in Tab.~\ref{tab:building}. Our method outperforms both the fully connected segmentation methods, as well as the active-contour based methods. Fig.~\ref{fig:buildingsample} presents typical samples, demonstrating the refinement that is obtained by the polygons over the initial segmentation mask.

The results for MoNu dataset are reported in Tab.~\ref{tab:Monu}. We outperform all baselines for both the Dice score and mean-IoU. Our algorithm also obtains better performances in comparison to the fully convolutional segmentation network with the same backbone Hardnet-85. The improvement from low resolution mask ($M$) to the output mask ($M_o$) is from 49.13\% IoU to 66.82\%. The full resolution FCN-Hardnet85 obtains a lower result of 66.06 IoU. Fig.~\ref{fig:Monu_seg} presents the output mask of our algorithm for samples from the test set. 

\begin{figure}[t]

    \centering
    \begin{tabular}{@{}c@{~~}c@{~~}c@{~~}c@{~~}}

    \includegraphics[width=0.22\linewidth]{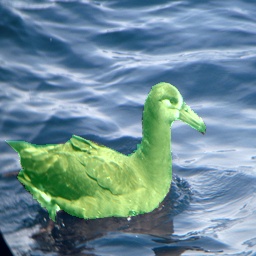} &
    \includegraphics[width=0.22\linewidth]{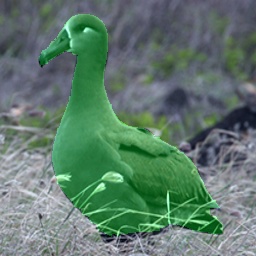} &
    \includegraphics[width=0.22\linewidth]{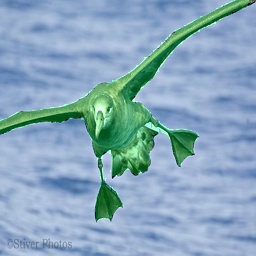} &
    \includegraphics[width=0.22\linewidth]{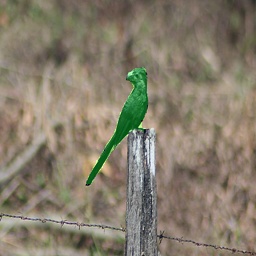} \\
    \end{tabular}
    \caption{Sample results of our method on the CUB dataset} 
    \label{fig:CUB}
\end{figure}

The results for CUB-200-2011 are reported in Tab.~\ref{tab:CUB}. We obtain better results than both FCN-Hardnet85 and Deep Contour on all metrics. Fig.~\ref{fig:CUB} present the output mask of our algorithm for samples from CUB-200-2011.

\begin{table}[t]
\centering
\begin{tabular}{@{}lccc@{}}
\toprule
Method & Dice & mIoU & FBound \\
\midrule
U-Net \cite{ronneberger2015u}  & 93.84 &  88.80 & 76.54 \\
Deep contour \cite{gur2019end} & 93.72 &  88.35 & 75.92\\
FCN-Hardnet85 & 94.91 & 90.51 & 79.73\\
Low res FCN-Hardnet85 ($M$) & 84.45 & 73.43 & 39.47\\
Ours ($M_o$) & {\bf95.11} & {\bf90.91} & {\bf79.86}\\
\bottomrule
\end{tabular}
\caption{Results for CUB dataset}
\label{tab:CUB}
\end{table}

\begin{figure}[t]
\centering
    \includegraphics[scale=0.425]{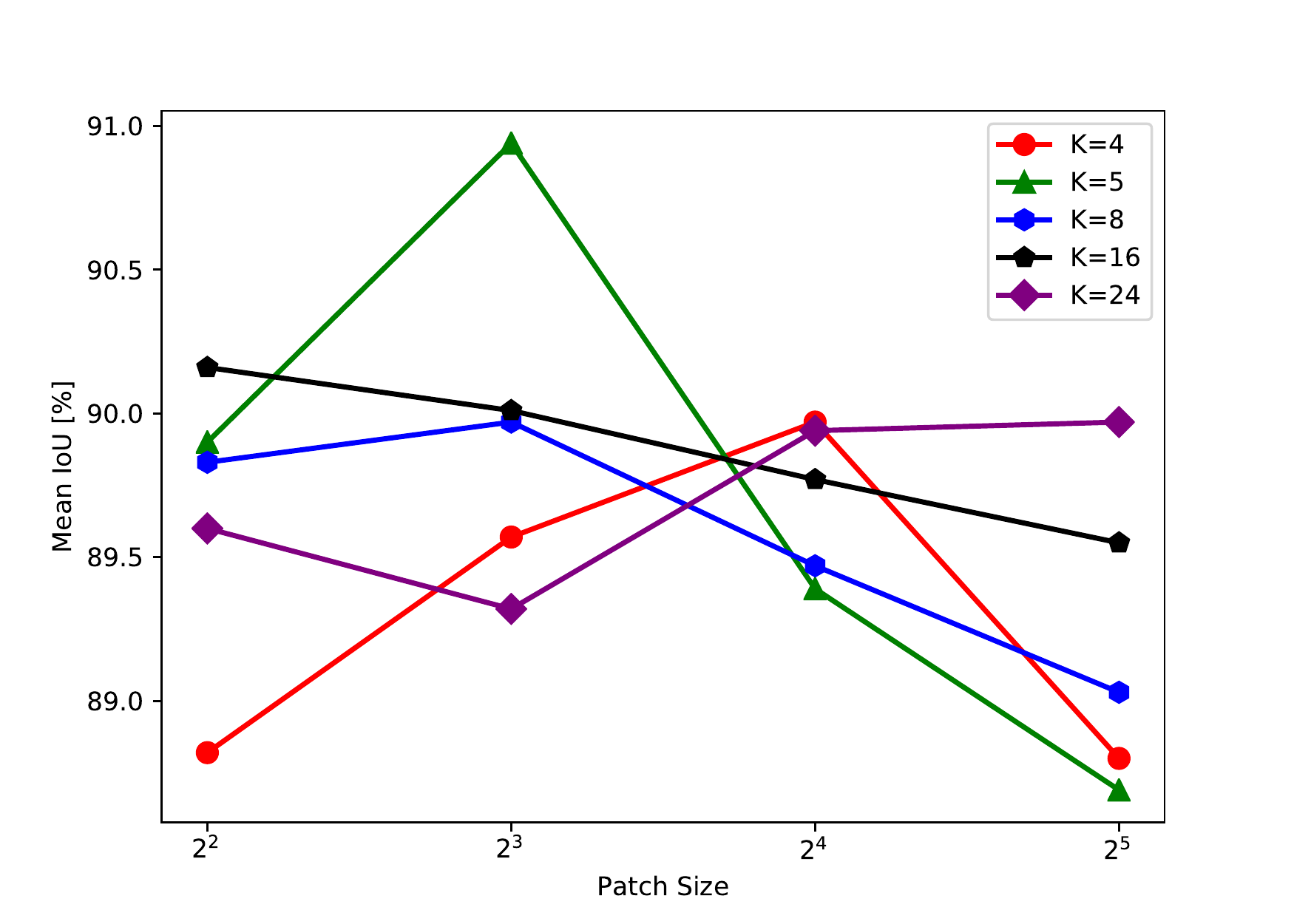}
    \caption{The effect on accuracy of varying the number of vertices to be one of [4, 5, 8, 16, 24] and the patch size to be in the set [4, 8, 16, 32]. Results are reported on the Vaihingen building dataset.}
    \label{fig:ablation}
\end{figure} 

\noindent{\bf Parameter sensitivity\quad} The proposed patch-wise polygon scheme adds two hyperparameters: the size of the patch that each polygon for each class is applied to and the number of vertices $k$ in each polygon. In order to study the effect of these parameters, we varied the number of vertices to be in the range of $[4, 5, 8, 16, 24]$ and the size of the patch to be one of $[4\times 4, 8\times 8, 16\times 16, 32\times 32]$.

Fig.~\ref{fig:ablation} shows our network performances for building segmentation, when varying the two parameters. As can be seen, for small patches, for instance, $4\times4$ or $8\times 8$, the number of nodes that optimizes performance is five. As may be expected, for larger patches, the method performs better with a higher number of vertices. 

We also checked our performances for different Hardnet backbones - HarDNnet39DS, HarDNnet68DS, HarDNnet68, HarDNnet85 (DS stands for for depth-wise-separable, see~\cite{chao2019hardnet}). The results appear in Tab.~\ref{tab:backbonebuilding} for the building dataset. As expected, the performance improves with the increase in the number of parameters in the network. It can also be observed that the low- resolution mask after upsampling $M$ and the refined mask $M_o$ have a performance gap of 4\% in terms of mean IoU, in favor of the latter, across all of the backbones.

\begin{table}[t]
\centering
\begin{tabular}{lccc}
\toprule
Backbone & Params[\#] & $M_o$ [mIoU] & $M$ [mIoU] \\
\midrule
HarDNet-39DS & 3.5M & 89.11 & 85.07  \\
HarDNet-68DS & 4.2M & 89.75 & 85.40 \\
HarDNet-68   & 17.6M & 90.50 & 85.60 \\
HarDNet-85   & 36.7M &  {\bf 90.92} & {\bf 85.98}\\
\bottomrule
\end{tabular}
\caption{Backbone comparison for the building segmentation task using various HarDNet variants~\cite{chao2019hardnet}. Showing the results of the upscaled gating signal $M$ and the final output $M_0$ that includes the polygons' boundaries.}
\label{tab:backbonebuilding}
\end{table}

\section{Discussion}
The ability to work with polygons allows our network to produce segmentation maps at a resolution that is limited only by machine precision. This can be seen in Fig.~\ref{fig:poly}(b), where the polygons have fractional coordinates. 
These polygons can be rasterized at any resolution.

The property of practically-infinite resolution may be of use in applications that require an adaptive resolution, depending on the object or its specific regions. An example is foreground segmentation for background replacement (virtual ``green screens''), in which hair regions require finer resolutions. Such scenarios would raise research questions about training networks at a resolution that is higher than the resolution of the ground truth data, and on the proper evaluation of the obtained results.

We note that while we do not enforce the polygons of nearby patches to be compatible at the edges, this happens naturally, as can be seen in Fig.~\ref{fig:poly}(b). Earlier during development, we have attempted to use loss terms that encourage such compatibility but did not observe an improvement.

Further inspection of the obtained polygons reveals that polygons with $k>4$ often employ overlapping vertices in order to match the square cell edges. In empty regions, polygons tend to become of zero-area by sticking to one of the boundaries. These behaviors emerge without being directly enforced. In some cases, phantom objects are detected by single polygons, as in the top right part of Fig.~\ref{fig:poly}(b). These polygons are removed by the gating process that multiplies with $M$, as can be seen in  Fig.~\ref{fig:poly}(c). We do not observe such polygons in the output images.

\section{Conclusions}
The pixel grid representation is commonly used by deep segmentation networks. In this work, we present an alternative approach that encodes the local segmentation maps as polygons. The method employs a neural renderer to backprop the rasterization process. A direct comparison to a method that employs the same backbone but without the polygonal representation reveals a significant improvement in performance on multiple benchmarks, despite using fewer training parameters. Moreover, our method also outperforms all other segmentation methods. 

\section*{Acknowledgment}
This project has received funding from the European Research Council (ERC) under the European
Unions Horizon 2020 research and innovation programme (grant ERC CoG 725974).

{\small
\bibliographystyle{ieee_fullname}
\bibliography{egbib}
}

\end{document}